\documentclass[letterpaper, 10pt, conference]{ieeeconf}  

\IEEEoverridecommandlockouts                              

\usepackage[T1]{fontenc}
\usepackage{amsmath} 
\usepackage{amssymb}  
\usepackage{xspace}
\usepackage{graphicx}
\usepackage{booktabs}
\usepackage{cite}
\usepackage{xcolor}
\usepackage{flushend}
\usepackage[font=small]{caption}
\usepackage{multirow}
\usepackage{etoolbox}
\usepackage[bookmarks=true]{hyperref}
\usepackage{bm}
\hypersetup{colorlinks,breaklinks,
linkcolor=[rgb]{0.5,0.,0.},
citecolor=[rgb]{0.000,0.427,0.173},
urlcolor=[rgb]{0.031,0.318,0.612}}

\usepackage{array}
\newcolumntype{L}[1]{>{\raggedright\let\newline\\\arraybackslash\hspace{0pt}}m{#1}}
\newcolumntype{C}[1]{>{\centering\let\newline\\\arraybackslash\hspace{0pt}}m{#1}}
\newcolumntype{R}[1]{>{\raggedleft\let\newline\\\arraybackslash\hspace{0pt}}m{#1}}

\newcommand{\papername}{Car-Studio}
\newcommand{\methodname}{\papername\xspace}

\makeatletter
\apptocmd{\@maketitle}{\centering \includegraphics[width=0.98\linewidth]{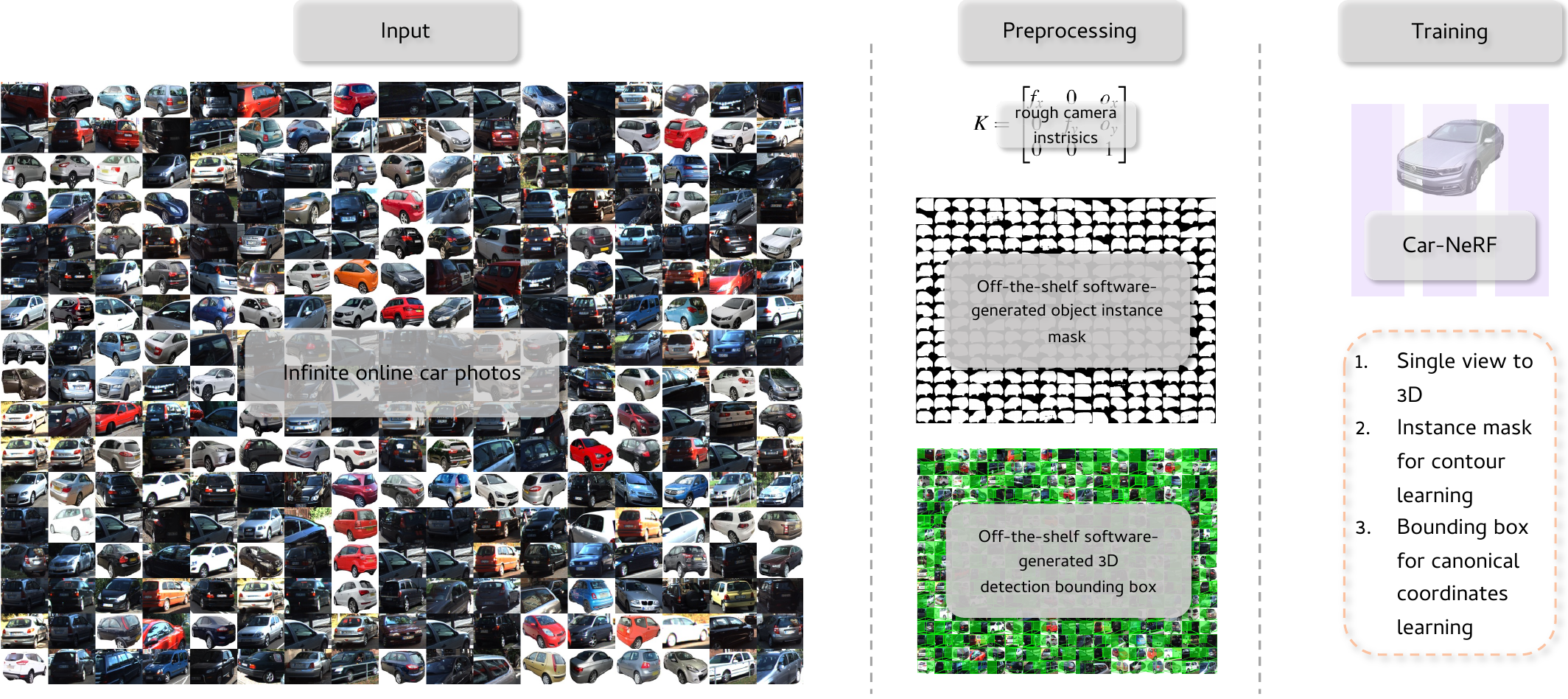}
\vspace{-0.5em}
\captionof{figure}{\textbf{\methodname: learns a General Car-NeRF from Infinite Car Images in the wild}. We extract car patches from unconstrained internet images using an off-the-shelf 2D object detector \cite{wu2019detectron2}. To preprocess the patches, we generate rough camera intrinsics, object instance masks, and 3D detection bounding boxes using a monocular 3D object detector \cite{dd3d} and a pixel-level segmenter \cite{SAM}. Our Car-NeRF model takes the preprocessed information as input and learns from a unified canonical car model with 3D bounding boxes and pixel-level masks to output a high-quality car image from any 3D viewpoint based on a single-view image.
\vspace{-0.7em} 
\label{fig:teaser}
}}{}{}
\makeatother

\begin{document}

\title{\LARGE \bf
\papername: Learning Car Radiance Fields from Single-View and Endless In-the-wild Images}

\author{Tianyu Liu$^{1}$, Hao Zhao$^{2}$, Yang Yu$^{3}$, Guyue Zhou$^{2}$, Ming Liu$^{1,3}$
\thanks{
$^{1}$ The Hong Kong University of Science and Technology, $^{2}$ AiR, Tsinghua University, $^{3}$The Hong Kong University of Science and Technology(Guangzhou)}
}

\maketitle
\thispagestyle{empty}
\pagestyle{empty}

\begin{abstract}

Compositional neural scene graph studies have shown that radiance fields can be an efficient tool in an editable autonomous driving simulator. However, previous studies learned within a sequence of autonomous driving datasets, resulting in unsatisfactory blurring when rotating the car in the simulator. In this letter, we propose a pipeline for learning unconstrained images and building a dataset from processed images. To meet the requirements of the simulator, which demands that the vehicle maintain clarity when the perspective changes and that the contour remains sharp from the background to avoid artifacts when editing, we design a radiation field of the vehicle, a crucial part of the urban scene foreground. Through experiments, we demonstrate that our model achieves competitive performance compared to baselines. Using the datasets built from in-the-wild images, our method gradually presents a controllable appearance editing function. We will release the dataset and code on \url{https://lty2226262.github.io/car-studio/} to facilitate further research in the field.

\end{abstract}

\section{INTRODUCTION}

NeRF\cite{Nerf} has emerged as a powerful art for generating novel views in both computer vision and computer graphics, and has recently garnered attention for its potential in autonomous driving\cite{autorf, snerf, PNF, SUDS, discoscene, neuralscenegraphs, gina3d, geosim, ners, StreetSurf, BlockNeRF, FEGR, urbangiraffe}. By modeling the color and density of every point within the relevant 3D space, NeRF can create highly realistic images of a scene from any viewpoint. This capability is particularly useful in the context of autonomous vehicle testing\cite{airsim, carla}, as it enables the replay of a scene multiple times with different viewpoints even though it's not recorded before. By doing so, NeRF can help improve the safety and reliability of autonomous vehicles by enabling more thorough testing and evaluation. 

To create a photorealistic and user-friendly autonomous driving testing platform based on recorded data, it is necessary first to reconstruct editable urban scenes. However, reconstructing editable urban scenes in the wild poses several challenges. One major challenge is the inconsistent movement over time between the background and foreground. Previous works, such as \cite{autorf, snerf, PNF, SUDS, discoscene, neuralscenegraphs, gina3d, geosim, ners, StreetSurf, FEGR}, have addressed this issue by decomposing the scene into background and foreground and learning foreground nodes based on categories, speed, distance, or a combination of these factors. In this letter, we focus on the challenge of reconstructing editable car instances and address it by parsing the foregrounds based on categories to achieve flexible editing.

Another challenge in reconstructing editable urban scenes is constrained views. Prior works discussed in the previous paragraph were trained on autonomous driving street datasets, where photos are captured from egocentric video streams that mostly show the front or rear of the car facing the camera. Consequently, these models may not perform well when the target is viewed from an unseen angle, such as the side view of a car. Some existing works have introduced pre-training generative branches \cite{discoscene, gina3d, urbangiraffe} to generate unseen views, but these indirect approaches do not address the fundamental problem of a shortage of side view data. To overcome the challenge of constrained views, we propose leveraging online photos captured in the wild, specifically infinite Internet images. However, using these images presents a new set of challenges, such as unknown illuminations, environments, camera intrinsic parameters, 3D poses, and dimensions. Therefore, we propose a novel toolchain that leverages in-the-wild 2D photos for favorable 3D neural radiance field training. Our approach aims to address the fundamental shortage of side view data and overcome the challenges of using online photos for 3D object reconstruction.

\begin{table*}[]
\vspace{1.5em}
\caption{Comparison of three dataset sources for CarPatch3D.} 
\label{tab:dataset}
\begin{center}
\begin{tabular}{ccccccc}
\toprule
Dataset &
  \begin{tabular}[c]{@{}c@{}}Background\end{tabular} &
  \begin{tabular}[c]{@{}c@{}}Unknown camera \\ intrinsics and poses\end{tabular} &
  \begin{tabular}[c]{@{}c@{}}Car orientaion\end{tabular} &
  \begin{tabular}[c]{@{}c@{}}Viewpoint diversity\end{tabular} &
  \begin{tabular}[c]{@{}c@{}}Filtered(details in \ref{sec:preprocessing}) \\ car patches number\end{tabular} &
  \begin{tabular}[c]{@{}c@{}}Filtered(details in \ref{sec:preprocessing}) \\ car instances number\end{tabular} \\
\midrule
KITTI-MOT & \checkmark   & \texttimes{} & Mostly front/back-view  & Single view + multi-view & 4481   & 383    \\
KITTI-DET & \checkmark   & \texttimes{} & Mostly front/back-view  & Single view only      & 4945   & 4945   \\
DVM-Cars  & \texttimes{} & \checkmark   & Many side-view & Single view + multi-view & 521275 & 201075 \\
\bottomrule
\end{tabular}
\end{center}
\vspace{-2.5em}
\end{table*}

Our toolchain, named \textbf{Car-Studio} (as depicted in Fig \ref{fig:teaser}), is designed to utilize in-the-wild car images to train a category-based NeRF \cite{Nerf} model for cars. To generate a car instances dataset, we use three in-the-wild datasets: KITTI dataset multi-object tracking track (KITTI-MOT)\cite{kitti}, KITTI dataset object detection track (KITTI-DET), and DVM-Cars\cite{dvm} dataset. We provide a comparison of these datasets in Tab \ref{tab:dataset}. While using in-the-wild data presents challenges such as messy objects with backgrounds, no camera parameters, unknown car poses, and single-view only instances without multi-view supervision, it also provides benefits.  For example, it offers different types of car instances with various orientations that supplement the urban benchmark such as \cite{kitti}, which lacks side-views.  

Based on our analysis of the combined datasets, we observed that the 2D car patches exhibit variations in size and possess boundaries that are challenging to discern, particularly in scenarios where the cars are in areas of low light or have hues that are similar to the surrounding environment. To overcome these challenges, we developed \textbf{Car-Nerf}, which leverages the anti-aliasing advantages of mip-NeRF \cite{mipnerf} to enable continuous scaling editing from far to near in simulation, and a segmentation mask that supervises the rendered accumulated radiance field weights for sharp contours. Additionally, we employ a canonical car-centric coordinate system to learn 3D spatial features from single-view only instances. Our proposed approach produces plausible rendering results in novel views and enables controllable spatial and appearance editing during scenes. 

Our contributions can be summarized as follows:

\begin{itemize}
    \item A curated dataset \textbf{CarPatch3D} of hundreds of thousands of 2D images and 3D spatial information. This dataset provides favorable information for training a category-based NeRF model for cars. Its availability enables the development of more effective urban NeRF foreground models.
    \item We developed \textbf{Car-NeRF}, which conforms to the characteristics of autonomous driving environments and achieves state-of-the-art performance in image reconstruction and novel view synthesis tasks.
    \item  We designed a pipeline called \textbf{Car-Studio} that can learn from single-view in-the-wild car images to generate 3D surrounding views and enable plausible controllable spatial and appearance editing.
\end{itemize}

\section{RELATED WORK}

{\bf Urban NeRF for autonoumous driving}. Urban NeRF has recently attracted considerable attention, with numerous methods and models developed to address the challenges posed by foreground objects, particularly cars. While some methods, such as BlockNerf \cite{BlockNeRF}, employs foreground masks to ignore the foreground objects and focus on large-scale background reconstruction, our work focuses on the foreground cars and their interactions with the environment. Existing methods \cite{neuralscenegraphs,snerf,PNF,SUDS,StreetSurf,DynamicViewSynthesis} use the same auto-decoder architecture as the seminal NeRF \cite{Nerf}. While these methods are effective for representing different instances, they lack the ability for zero-shot learning and controllable appearance editing. Recent approaches\cite{discoscene, urbangiraffe, gina3d,nerfdiff,3DiM,Zero1to3,deceptivenerf,Magic123} have utilized generative models to learn foreground models. In contrast, our approach adopts the decoder-encoder architecture used in previous works such as PixelNerf \cite{PixelNeRF} and AutoRF \cite{autorf}, without relying on explicit generators. Moreover, our approach extends the processing of images to include pose-free and even camera intrinsics-free in-the-wild images and introduces a controllable appearance editing approach to enable more flexible editing capabilities.

{\bf Single view to 3D NeRF}. Inferring 3D shapes from 2D images is a long-standing challenge in computer vision and graphics. Existing methods present good results with different approaches. Deformation-based methods\cite{CMR, pixel2mesh,deformation,nvdiffrec} learn the 3D shape by mesh deformation based on a primitive mesh. Unsup3D\cite{unsup3d}  assumes symmetric and deformable shapes to learn 3D shapes. Mesh R-CNN \cite{meshrcnn} predicts the 3D mesh by a 3D voxel prediction followed by a mesh refine branch. PiFu \cite{pifu} learns the occupancy field in 3D directly by the occupancy  implicit functions.  Some works learn shape codes for different types of object priors like CAD model\cite{3D-RCNN, CADSim}, SDF-based Database\cite{singleshotsr}, activation code library\cite{ObjectCompositinalNeRF}.  Recently, single-view to 3D NeRF has emerged as a promising approach to tackling this problem, with impressive results achieved by methods such as MVSNerf \cite{MVSNerf} and PixelNerf \cite{PixelNeRF} using multi-view supervision, obtaining multi-view images is often more expensive and impractical compared to single-view images, especially in real-world scenarios. In contrast, our proposed method can process both multi-view and single-view image input, making it more versatile and practical for a wider range of applications. Recent works have explored the use of transformer-based architectures rather than MLPs to develop methods for view synthesis from a single input image, offering more fine-grained control over the reconstruction process \cite{visionNerf, SRT}. CodeNeRF \cite{codenerf} initializes the implicit representation with a pre-trained neural network and optimizes it using a differentiable renderer. Our proposed method differs in that ours involve two stages: an encoder that extracts features from the input image and uses them to initialize the implicit representation, followed by a similar optimization process as CodeNeRF to refine it further. This enables our model to extract individual instance latent codes from images with backgrounds, while CodeNeRF is restricted to test optimization only.

\section{Methodology}
\subsection{\methodname Pipeline Overview}\label{sec:overview}

We propose \textbf{\methodname}, a three-phase pipeline designed to learn a general car neural rendering model from online car photos, as shown in Fig. \ref{fig:teaser}. In the first phase, \textbf{Input}, we use a pre-trained 2D detector \cite{wu2019detectron2} to identify patches of interest. We filter out patches that are too small or whose classification results indicate a non-vehicle object to ensure high-quality data. In the subsequent \textbf{Preprocessing} phase, we refine the data further to prepare it for use in our model. We describe the preprocessing phase in detail in Section \ref{sec:preprocessing}. In the third phase, \textbf{Training}, we use our \textbf{Car-NeRF} model, discussed in Section \ref{sec:car-nerf}, to learn from the preprocessed data.

\subsection{Preprocessing}\label{sec:preprocessing}

The NeRF model \cite{Nerf} requires images, camera parameters, and camera poses as input. Obtaining accurate camera parameters from a single image can be challenging when there is no known size calibration target in the field of view. Therefore, we use a rough focal length based on \cite{kitti} and the optical center equal to half of the image shape size, and pass them to a 3D detection model \cite{dd3d} to solve for the camera pose.  To ensure high-quality patches, we apply a filtering process that consists of the following steps:

\begin{enumerate}
\item We filter out images with a low intersection ratio between the 2D detector output and the minimum 2D envelope box for reprojecting the 3D box, as this indicates inconsistencies between the outputs of the 2D and 3D detectors.
\item We filter out images with a low pixel count ratio between the result of the 2D segmentor Segment Anything \cite{SAM} pixel-level segmentation and the 2D detector's result, which indicates high occlusion rate.
\item We filter out images with a low prediction confidence from the 2D detector, 3D detector, and 2D segmentor.
\item We manually filter out 3D prediction results that are inconsistent with the image.
\end{enumerate}

We carefully followed the aforementioned steps to ensure high-quality preprocessing of the data, making it suitable for use in our pipeline. This effort resulted in a new dataset named \textbf{CarPatch3D}. CarPatch3D comprises a set of car patches, selected using the semi-automatic method described above, along with binary masks generated by the SAM \cite{SAM}, roughly estimated camera parameters, and 3D position and orientation of the vehicles generated by DD3D \cite{dd3d}. This rich set of data makes CarPatch3D highly suitable for NeRF training of cars in autonomous driving.

\subsection{Car-NeRF Model}\label{sec:car-nerf}

\begin{figure*}
\vspace{1.5em}
\centering
\includegraphics[width=0.95\linewidth]{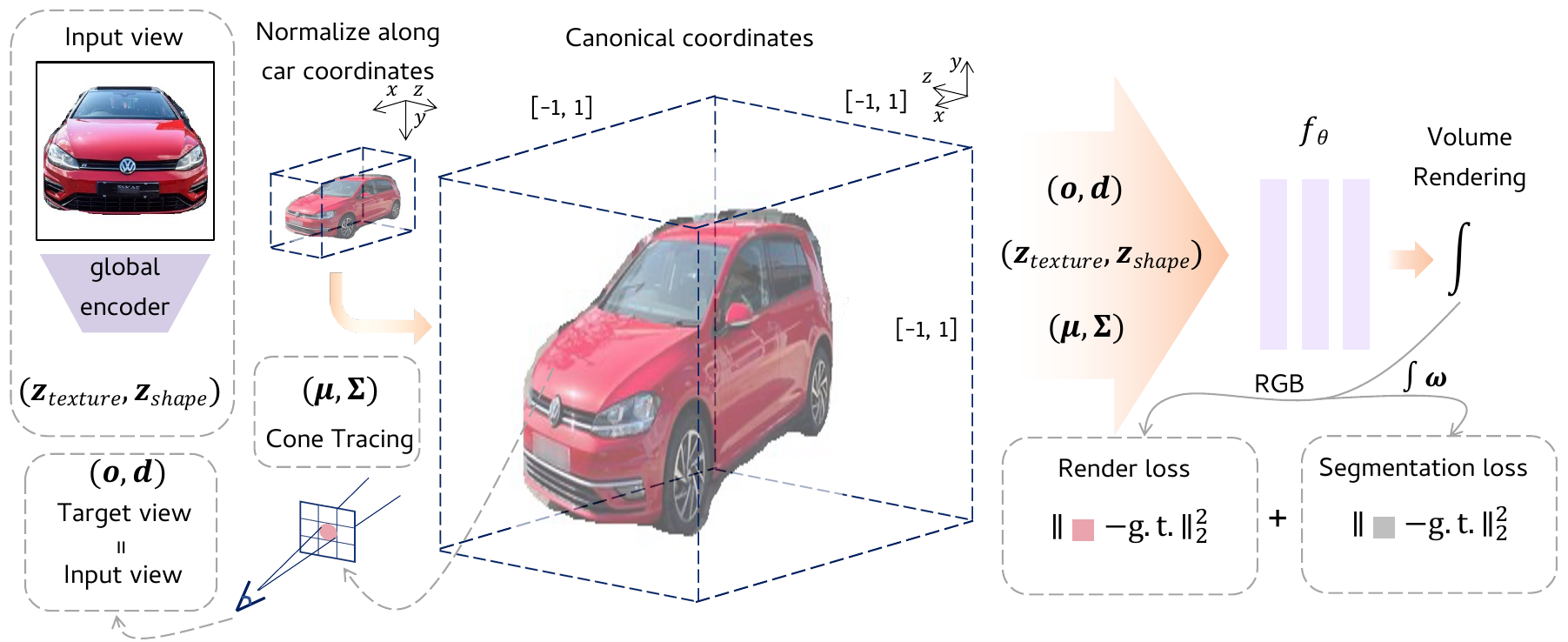}
\caption{The Car-NeRF architecture. We use a global encoder to obtain texture ($\bm{z}_{texture}$) and shape ($\bm{z}_{shape}$) latent vectors from the input image. We then scale the world coordinates into object-centric normalized canonical coordinates. Cone tracing is performed at the input view's position $\bm{o}$ along direction $\bm{d}$ to obtain the mean distance $\bm{\mu}$ and variance $\mathbf{\Sigma}$ along the axis of the cone, which are passed to our model  $f_\theta$. The model outputs the rendered image's RGB value, which is used to calculate render loss, while the segmentation losses are calculated using the output accumulated weights along axis $\int \bm{\omega}$.\vspace{-16pt}}\label{fig:architecrure}
\vspace{-0.5em}
\end{figure*}

The architecture of our model is shown in Fig \ref{fig:architecrure}. We learn the global latent codes $\bm{z}$ using a global encoder, which is a ResNet-34\cite{resnet} pre-trained on ImageNet \cite{imagenet}. We then decouple the global latent codes into a shape component $\bm{z}_{shape}$ and a texture component $\bm{z}_{texture}$ using two one-layer independent MLPs.

\textbf{Normalization to car-centric canonical coordinate}.To learn 3D shapes from single-view image supervision, we convert the camera-centric coordinates to canonical coordinates centered at the car's centroid and oriented along its principal axes. We obtain the dimensions of the car (length $l$, height $h$, and width $w$) from the preprocessing 3D detection module's result. The scaled [-1, 1] camera pose with respect to the camera coordinate $T^{scaled}_{cam}$ is given by:
\begin{equation}
    T^{scaled}_{cam} = ([\frac{2}{l}, \frac{2}{h}, \frac{2}{w}, 1]^T \mathbf{1}_{1\times4}) \circ \mathbf{I}_{4\times4}
\end{equation}, where $\circ$ is the Hadamard product.

The camera pose with respect to the canonical coordinate system is obtained by the scaled camera pose with respect to the camera coordinate system ($T_{cam}^{scaled}$), and the car pose with respect to the camera coordinate system ($T_{cam}^{car}$). The resulting camera pose with respect to the canonical coordinate system $T^{cam}_{canon}$ is given by:
\begin{equation}
T^{cam}_{canon} = T^{scaled}_{cam}(T_{cam}^{car})^{-1} T^{scaled}_{canon}
\end{equation}, where $T^{scaled}_{canon}$ is scaling matrices that map the  canonical coordinates to camera-centric scaled coordinates whose diagonal values are (1, -1, -1, 1) according to Fig \ref{fig:architecrure}.

\textbf{Cone tracing and IPE}. In our architecture, the encoder takes Internet images processed by a 2D detector as input. Since the patch size of the input images can vary greatly, the region that each pixel represents can also vary greatly. To address this, we use cone tracing instead of ray tracing for a continuous scaling that provides anti-aliasing \cite{mipnerf}. A cone at apex $\bm{o}$ with direction $\bm{d}$ is represented by a multivariate Gaussian distribution with mean $\bm{\mu}(\bm{o}, \bm{d})$ and variance $\mathbf{\Sigma}(\bm{d})$.

To build a continuous scaling along the axis of cones, we use the integrated positional encoding (IPE) \cite{mipnerf} instead of positional encoding (PE) \cite{Nerf}. The IPE incorporates the multivariate Gaussian representation into the encoding process, resulting in improved performance. The IPE of positions is denoted by $\gamma_{pos}$, while the IPE of directions is denoted by $\gamma_{dir}$.

\textbf{Sampling and model architecture}. We adopt a coarse-to-fine pixel sampling strategy and a shared model architecture similar to mip-NeRF \cite{mipnerf}. The model $f_\theta$ consists of a shape component $f_{\theta, shape}$ and a texture component $f_{\theta, texture}$. To embed the latent vectors into the model, we introduce the following modifications:
\begin{equation}
\mathbf{f}_{out}, \sigma = f_{\theta, shape} (\gamma_{pos}(\bm{\mu}(\bm{o}, \bm{d}), \mathbf{\Sigma}(\bm{d})) + \bm{z}_{shape} )
\end{equation}

Here, $\bm{z}_{shape}$ is the shape latent vector, $\sigma$ is the density, and $\mathbf{f}_{out}$ is the output feature of the shape component. The RGB color, denoted by $c$, can be calculated using the texture latent vector $\bm{z}_{texture}$ as follows:
\begin{equation}
\textup{RGB} = c = f_{\theta, texture}(\mathbf{f}_{out} + \gamma_{dir}(\bm{d}) + \bm{z}_{texture})
\end{equation}

The accumulated weights $\int \bm{\omega}$ from near $t_n$ to far $t_f$ along the axis $\bm{r} = \bm{o} + t\bm{d}$ of the cone is given by:
\begin{equation}
    \smallint \bm{\omega} (\bm{r}) = \int_{t_n}^{t_f}\exp(-\int_{t_n}^t\sigma(\bm{r}(s)ds))\sigma(\bm{r}(t))dt
\end{equation}

\textbf{Volume Rendering and Loss Functions}. We employ the volume rendering technique described in \cite{Nerf} to compute the estimated rendered color $\hat{\textup{RGB}}$ and the estimated accumulated weights $\hat{\int \bm{\omega}}$ along the cone axis defined by the apex $\bm{o}$ and the direction $\bm{d}$ at the given pixel location. At this pixel location, the ground truth color is denoted as $\textup{RGB}$, and the corresponding binary mask obtained from the 2D segmentor described in Section \ref{sec:preprocessing} is denoted as $\alpha$. The photometric L2 render loss $\mathcal{L}_r$ is defined as follows:
\begin{equation}
    \mathcal{L}_{r} = \alpha(|| \hat{\textup{RGB}}_{f} - \textup{RGB}||_2^2 + \lambda_{c}|| \hat{\textup{RGB}}_{c} - \textup{RGB}||_2^2)
\end{equation}

Here, the estimated rendered color obtained from fine sampling is denoted as $\hat{\textup{RGB}}_{f}$, while that obtained from coarse sampling is denoted as $\hat{\textup{RGB}}_{c}$. The hyperparameter $\lambda_{coarse}$ is used to balance the loss between two sampling stages. To prevent our model from learning the background color or shapes, we incorporate the binary mask $\alpha$ into the loss function. Similarly, the segmentation loss $\mathcal{L}_s$ is:
\begin{equation}
    \mathcal{L}_s = ||\hat{\smallint \bm{\omega}}_f - \alpha||^2_2 + \lambda_c||\hat{\smallint \bm{\omega}}_c - \alpha||^2_2
\end{equation}

The total loss is weighted by a segmentation balancing coefficient $\lambda_s$:
\begin{equation}
    \mathcal{L} = \mathcal{L}_{r} + \lambda_s \mathcal{L}_{s}
\end{equation}

\section{EXPERIMENTS}

We construct our datasets \textbf{CarPatch3D} using three different sources: the KITTI multi-object tracking (KITTI-MOT) and object detection (KITTI-DET) tracks \cite{kitti}, and the DVM-Cars dataset \cite{dvm}. The characteristics of each dataset are shown in Table \ref{tab:dataset}. Our work addresses three challenging tasks: single-view supervision for zero-shot learning, multi-view supervision for novel scene synthesis, and patch reconstruction with test-time optimization. Furthermore, we demonstrate Car-Studio's ability to perform controllable scene editing. 

\textbf{Baselines}. We evaluate our proposed method against the following baselines: CodeNeRF \cite{codenerf}, PixelNeRF \cite{PixelNeRF}, and AutoRF \cite{autorf}. Note that CodeNeRF does not support zero-shot learning, and therefore, we only use it for the multi-view supervision few-shot learning task. Moreover, to ensure a fair comparison since our problem is category-specific, we convert PixelNeRF to the canonical coordinate system.

\textbf{Test-time Optimization}. For the auto-decoder model \cite{codenerf}, we optimize the latent vectors directly in the test dataset. However, for the encoder-decoder model \cite{PixelNeRF, autorf} and our model, we first obtain the latent vectors using the encoder and then perform joint optimization of the latent vectors and decoder during the test-time optimization stage.

\textbf{Runtime Environment Details}.  All experiments are conducted on an Nvidia RTX 3090 Ti graphics card. We optimize all NeRF models for 500,000 steps with a pixel sampler with a batch size of 3,072, which takes approximately two days to complete. Our pipeline is built upon NerfStudio \cite{nerfstudio}. We also reproduce the CodeNeRF \cite{codenerf}, Canonical PixelNeRF-ResNet\cite{PixelNeRF}, Canonical PixelNeRF-MLP with an MLP backbone, and AutoRF\cite{autorf} based on a community implementation\cite{autorf-pytorch} for reusing the data loader. We use the RAdam optimizer \cite{radam} with hyperparameters $\beta_1 = 0.9$, $\beta_2 = 0.999$, $\textup{lr} = 10^{-3}$, and $\epsilon = 10^{-8}$. We use the exponential decay scheduler with $\textup{lr}_{\textup{final}} = 10^{-4}$ and $\textup{step}_{\max} = 200,000$. All other configurations follow the original setups.

\begin{figure}
\centering
\vspace{0.5em}
\includegraphics[width=0.5\linewidth]{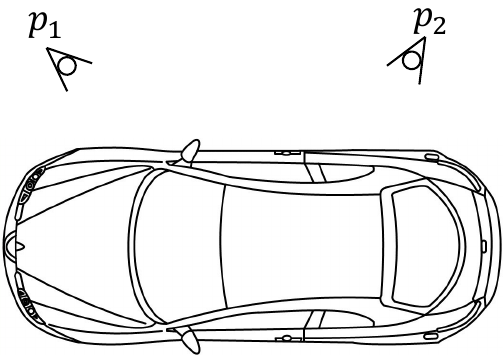}
\caption{Illustration of single-view and multi-view supervision.}\label{fig:single-view}
\vspace{-0.5em}
\end{figure}

\textbf{Evaluation and Metrics}. We evaluate the models on novel view synthesis. In Figure \ref{fig:single-view}, suppose we have two images of the same car instance captured at different poses $p_1$ and $p_2$. If we feed the image captured at $p_1$ to the encoder to obtain the latents, we can then render the image at the view $p_1$ to compute the loss between the rendered image and the ground truth image captured at $p_1$.This approach is called single-view supervision because it relies solely on information from a single view.If we were to use the image captured at $p_1$ and compute the loss using the rendered image at $p_2$, it would be considered multi-view supervision. The quantitative results are PSNR, SSIM \cite{ssim}, and LPIPS \cite{lpips}.

\subsection{Single-view Supervision for Zero-shot Learning} \label{sec:single-view}

\begin{figure*}[t]
\vspace{1.5em}
\includegraphics[width=0.95\linewidth]{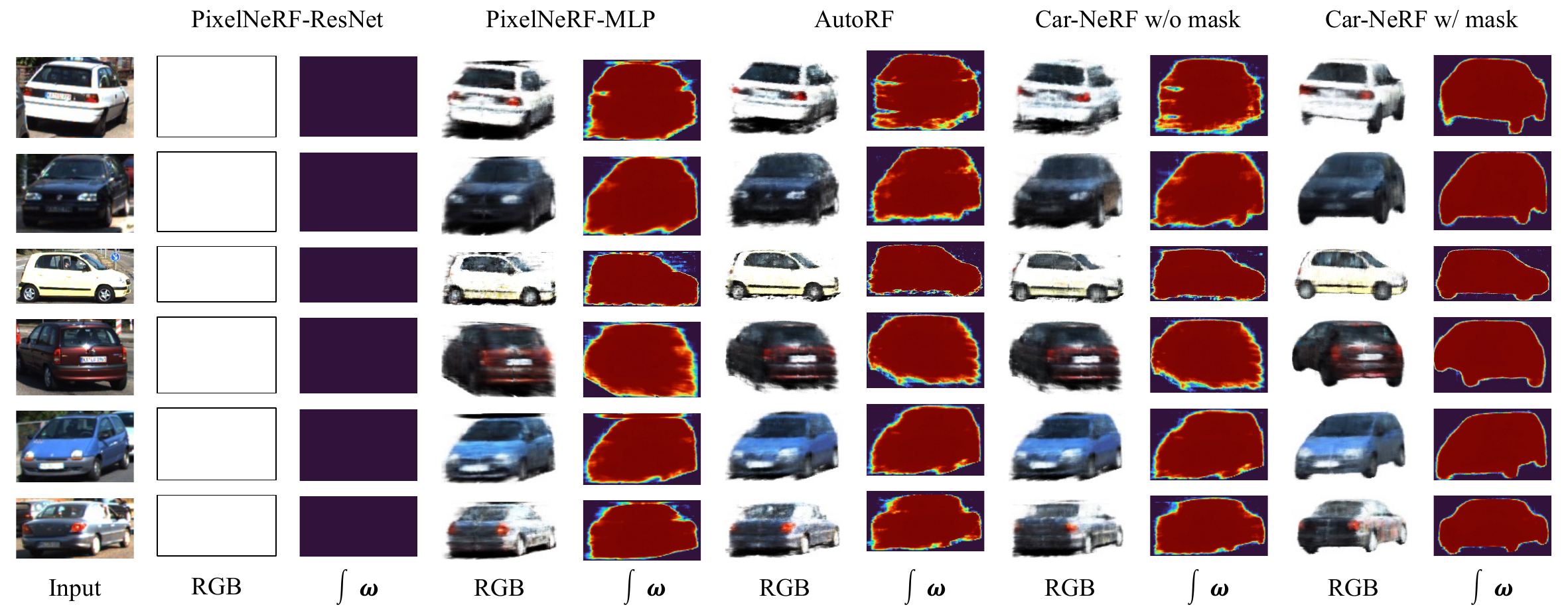}
\caption{Comparison of RGB and accumulated weights $\int \bm{\omega}$ rendering, using a zero-shot learning paradigm with all instances unseen in the training set.  The figure depicting accumulated density uses a Turbo color scheme, where dark blue represents low values (down to 0) and dark red represents high values (up to 1). \vspace{-1.4em}}\label{fig:zero_shot_qualitative}
\vspace{-0.6em}
\end{figure*}

\begin{table}[t]
\centering
\caption{Performance comparison of different models for single-view supervision zero-shot learning. The best result is highlighted in {\color{red}red}, and the second best result is highlighted in {\color{blue}blue}. } 
\label{tab:single_view_zero_shot}
\begin{tabular}{cccc}
\toprule
Method             & PSNR$\uparrow $       & SSIM$\uparrow $        & LPIPS$\downarrow$      \\
\midrule
PixelNeRF-ResNet   & 11.61               & 0.1948               & 0.7375               \\
PixelNeRF-MLP      & 17.87               & 0.5103               & 0.4014               \\
AutoRF             & {\color{blue}20.75} & 0.5903               & 0.3466               \\
\midrule
Car-NeRF(w/o mask) & {\color{red}20.93}  & {\color{blue}0.6020} & {\color{blue}0.3396} \\
Car-NeRF(w/ mask)  & 20.23               & {\color{red}0.6209}  & {\color{red}0.3284} \\
\bottomrule
\end{tabular}
\vspace{-1.5em}
\end{table}
In Figure \ref{fig:zero_shot_qualitative}, we present qualitative comparisons with the baselines. In this experiment, all testing instances are unseen in the training set, and no test-time optimization is performed. The experiment uses a uniform 9:1 train-test split on all valid KITTI-DET-derived patches. We observe that the default architecture of PixelNeRF-ResNet degenerates to an empty output. Incorporating the segmentation loss $\mathcal{L}_s$ into our approach results in sharper contours compared to the baselines. Additionally, our architecture design reduces the presence of floaters around the target car.

Table \ref{tab:single_view_zero_shot} presents the quantitative results of single-view supervision zero-shot novel view synthesis. PixelNeRF \cite{PixelNeRF} uses a ResNet-based backbone by default. However, during the training process, PixelNeRF-ResNet encountered degradation. As a result, we also compare the performance of an MLP-based version of PixelNeRF. The results show that Car-NeRF with mask segmentation loss achieves the best SSIM and LPIPS performance, while Car-NeRF without mask segmentation loss performs best in terms of PSNR. 

The quantitative results suggest that incorporating mask segmentation loss can improve the quality of rendered images in terms of structural similarity and perceptual quality. This finding is consistent with the intuition of using mask segmentation to guide the training of neural rendering models. However, it is important to note that Car-NeRF without mask segmentation loss performs better in terms of peak signal-to-noise ratio. This indicates that relying solely on PSNR may not provide a comprehensive evaluation of image quality. Therefore, it may be necessary to consider various metrics, such as structural similarity and perceptual quality, in addition to PSNR, to obtain a more complete assessment of the performance of neural rendering models.

\subsection{Muti-view Supervision Novel View Synthesis}

\begin{table}[t]
\centering
\vspace{0.5em}
\caption{Comparison of performance across models trained with multi-view supervision and single-view supervision. The best result is highlighted in {\color{red}red}. } 
\label{tab:multi_view}
\begin{tabular}{cccc}
\toprule
Method             & PSNR$\uparrow $       & SSIM$\uparrow $        & LPIPS$\downarrow$      \\
\midrule
PixelNeRF-MLP      & 17.16               &      0.4715          & 0.4343              \\
AutoRF & 17.84 & 0.5212 & 0.3852 \\
CodeNeRF & 19.01 & 0.5737 & 0.3299 \\
\midrule
Car-NeRF(multi-view) &17.40 & 0.4965 & 0.4075 \\
Car-NeRF(single-view)  & \color{red}20.12 & \color{red}0.6251 & \color{red}0.2825\\
\bottomrule
\end{tabular}
\vspace{-1.5em}
\end{table} 

In this experiment, we compared our method with other approaches to assess the effectiveness of single-view supervision versus multi-view supervision. While the baselines in Section \ref{sec:single-view} can obtain latent vectors by inputting an arbitrary image to the encoder, the auto-decoder architecture does not have this ability. To provide a fair comparison with these approaches, we use a more comprehensive dataset with instance labels across frames, which requires additional labor but provides more ideal evaluation conditions for multi-view supervision approaches. 

We train the models on patches derived from the KITTI-MOT dataset, using the cross-frame instance label as the ground truth for multi-view supervision. To ensure a fair comparison, we only consider instances that have multiple images available for training. This is because CodeNeRF \cite{codenerf} uses global latent vectors as input, and in the case where an instance is not seen in the training set, using an average initialization without any instance-specific information may introduce potential unfairness to CodeNeRF.
Similar to the experiment in Section \ref{sec:single-view}, we do not apply any test-time optimization to the models in this experiment.

\begin{figure}[t]
\centering
\includegraphics[width=0.98 \linewidth]{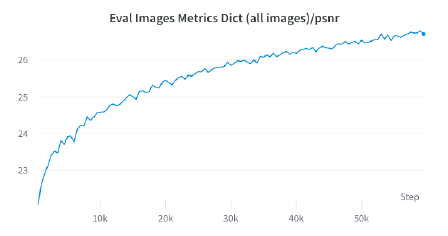}
\caption{Learning curve for test-time optimization \vspace{-1.4em}}\label{fig:psnr}
\vspace{-0.6em}
\end{figure}

\begin{figure}[t]
\centering
\vspace{1.5em}
\includegraphics[width=1.0 \linewidth]{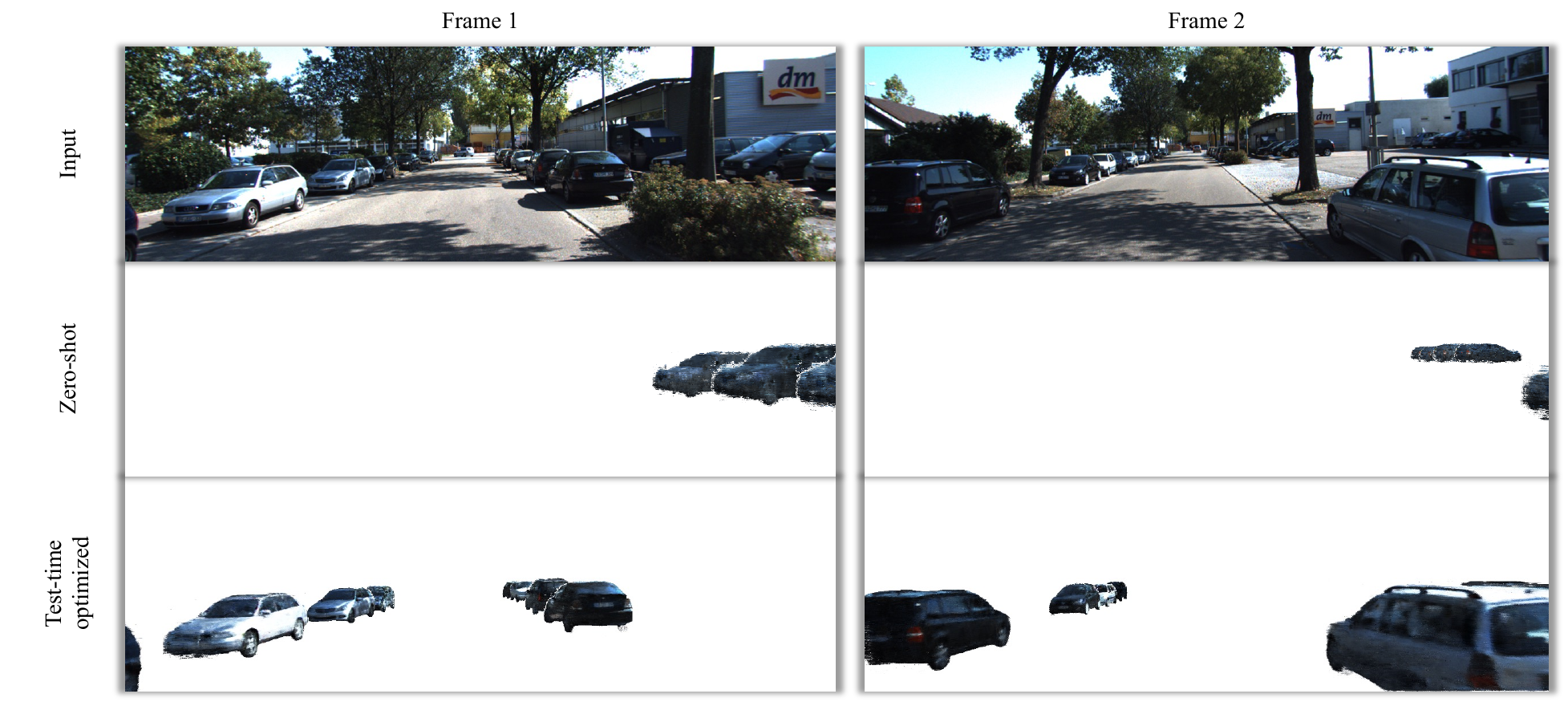}
\caption{Qualitative ablation study for test-time optimization. The third row shows instances with test-time optimization, achieving higher image quality compared to the second row without test-time optimization. Details on \href{http://lty2226262.github.io/car-studio/}{project page}.
\vspace{-1.4em}}\label{fig:test_time_optimization}
\vspace{-0.6em}
\end{figure}

\begin{figure}[t]
\centering
\includegraphics[width=0.9 \linewidth]{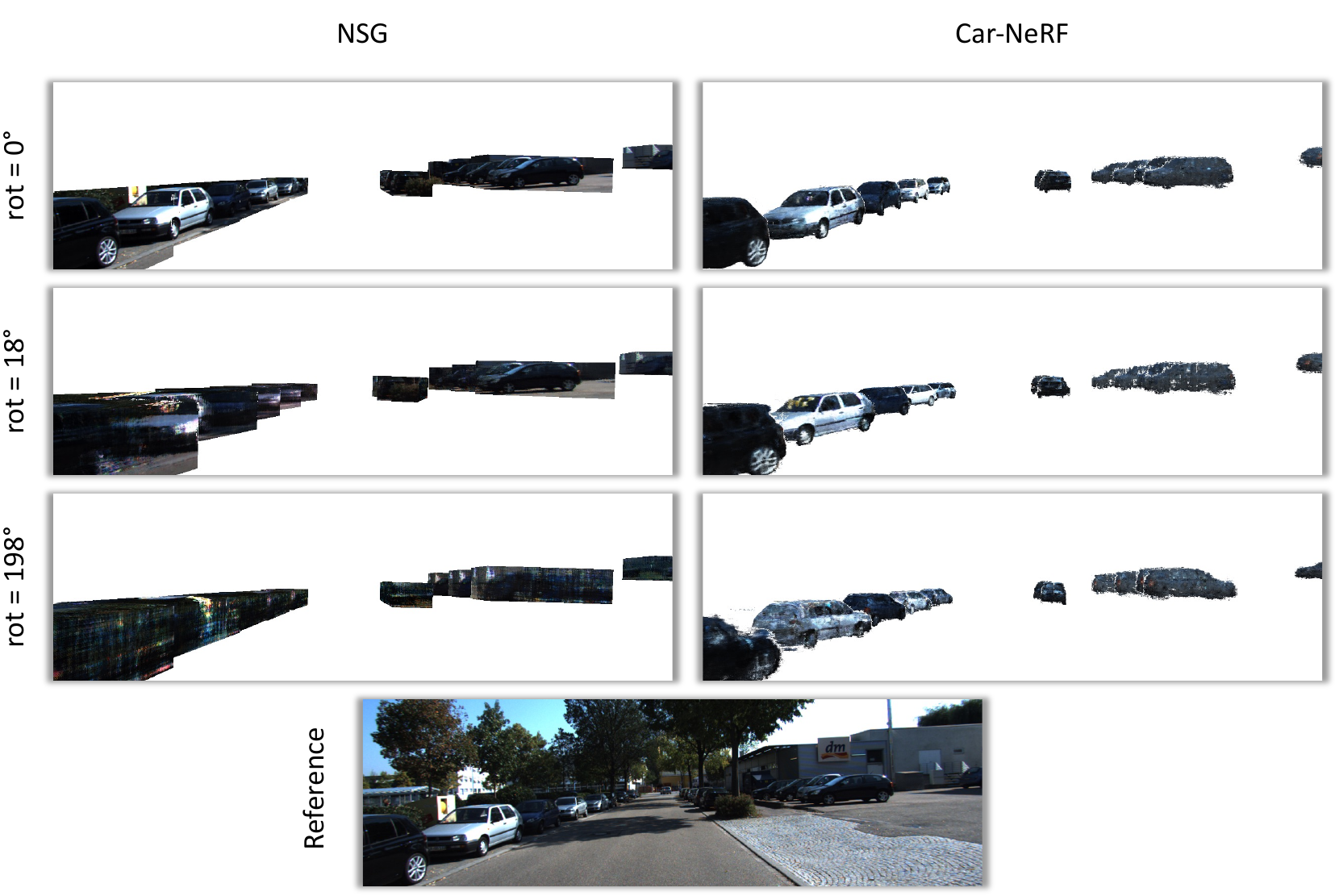}
\caption{Unseen view synthesis comparison between NSG \cite{neuralscenegraphs} and our method. The left column shows NSG's results, while the right column shows our results. Our method performs better than NSG in generating sharp and realistic images, even when a large rotation angle is applied. Specifically, while the images generated by NSG become blurry when an 18-degree rotation angle is applied, our method is still able to generate better quality images. When a larger rotation angle is applied, NSG fails to generate an image of the vehicle, whereas our method can still synthesize unseen views with higher fidelity. Details on \href{http://lty2226262.github.io/car-studio/}{project page}.
\vspace{-1.4em}}\label{fig:rot}
\vspace{-0.6em}
\end{figure}

The comparison results are presented in Table \ref{tab:multi_view}, which demonstrate that our Car-NeRF model achieves competitive performance with PixelNeRF-MLP \cite{PixelNeRF} and AutoRF \cite{autorf} in the multi-view supervision setting. Notably, our method that employs only single-view supervision outperforms all of the multi-view supervision methods. This observation could be attributed to the fact that the rough estimation of camera parameters introduces more inconsistent noise to the multi-view supervision framework. In contrast, using single-view supervision avoids this inconsistency by relying on a single viewpoint for training. Furthermore, CodeNeRF outperforms the encoder-decoder architecture approaches (all methods except CodeNeRF in this experiment) in the multi-view setting. This may be because encoder-decoder models use a global encoder to extract the latent vectors, which are optimized across all instances, while auto-decoder models optimize the latent vectors instance-specifically. Therefore, we further investigate the potential benefits of instance-specific optimization.

\subsection{Foreground Reconstruction with Test-Time Optimization}

To further leverage the advantages of independently optimizing the latent vector, akin to auto-decoder approaches, we implement a test-time optimization on the latent vector decoupled from the encoder. In the first stage, we train the Car-NeRF model using the KITTI-DET-derived dataset. We then detach the latent code from the encoder and proceed to the second stage of training. In the second stage, we perform a test-time optimization on the Car-NeRF's field and the latent codes using the KITTI-MOT-derived dataset. Figure \ref{fig:psnr} shows the quantitative learning curves for test-time optimization. Figure \ref{fig:test_time_optimization} presents qualitative comparisons between the results with and without test-time optimization. We also compare our approach with a no-prior method, taking NSG \cite{neuralscenegraphs} as an example in Figure \ref{fig:rot}. These comparisons demonstrate that our Car-NeRF model is scalable for test-time optimization while retaining the ability to synthesize unseen views.

\begin{figure}[t]
\centering
\vspace{1.5em}
\includegraphics[width=0.9 \linewidth]{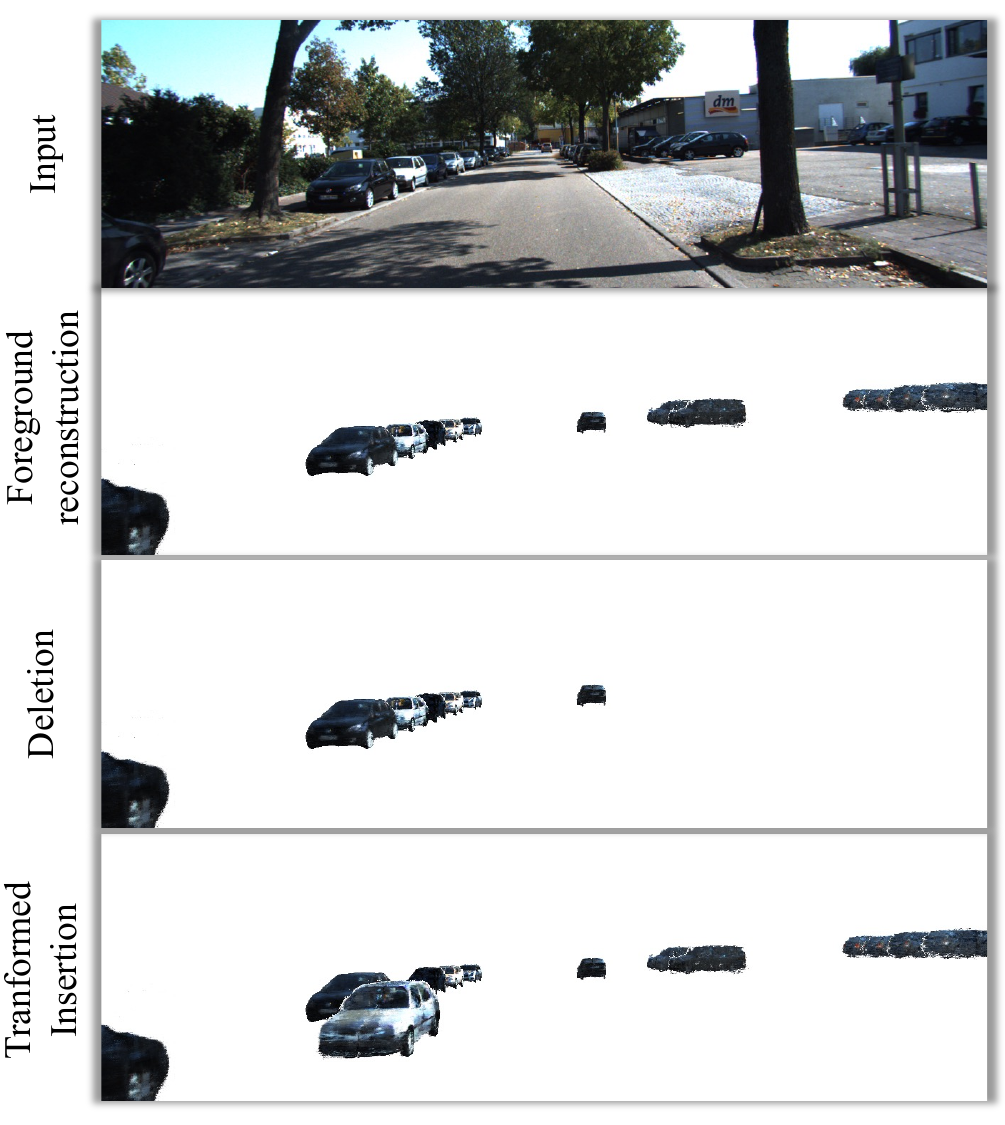}
\caption{Autonomous driving urban scene car instances editing showcases. Details on \href{http://lty2226262.github.io/car-studio/}{project page}.
\vspace{-1.4em}}\label{fig:manipulation}
\vspace{-0.6em}
\end{figure}

\subsection{Applications for Autonomous Driving}

Our new pipeline enables foreground editing in an autonomous driving simulator, including instance insertion, spatial transformation, deletion, and replacement of instances, as demonstrated in Figure \ref{fig:manipulation} and similar to NSG \cite{neuralscenegraphs}. However, our approach performs better on unseen views, as shown in Figure \ref{fig:rot}.

\begin{figure}[t]
\centering
\vspace{1.5em}
\includegraphics[width=0.9 \linewidth]{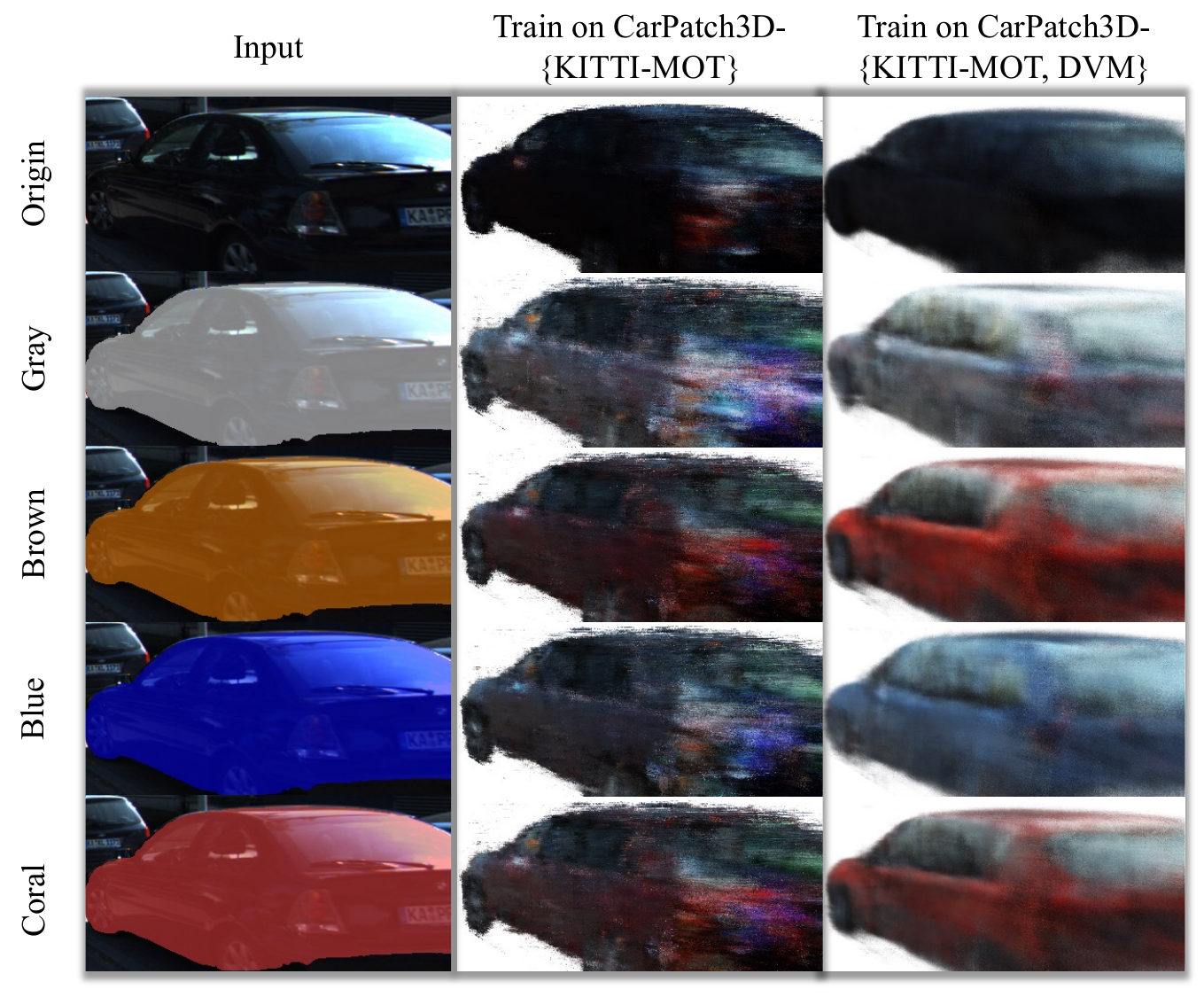}
\caption{Demonstration of appearance editing using various synthesis inputs for our model.
\vspace{-1.4em}}\label{fig:appearance}
\vspace{-0.6em}
\end{figure}

Furthermore, our model's ability to use larger datasets enables new applications.  We can now controllably and separately edit the appearance of objects without modifying their geometry features. This is demonstrated in Figure \ref{fig:appearance}, where a transparent color mask is applied to the original patch to achieve the desired appearance. The figure shows that our model enables controllable appearance editing, and this ability improves as the dataset scale increases.

\section{Conclusion}

In this paper, we presented \methodname, a pipeline for learning car's neural radiance fields from in-the-wild 2D images. Our pipeline includes a data processing component and a neural radiance field component.

We collected CarPatch3D, a dataset that provides multi-perspective images of cars by processing images from existing datasets with different design purposes to address the gap in existing autonomous driving datasets that often provide front or rear views of cars.

Our designed Car-NeRF, a canonical, cone-tracing neural radiance field with extra segmentation loss, enables our model to learn from single-view supervision, anti-aliasing, and sharp contours, leading to competitive performance in rendering realistic images of car objects.

Moreover, the trained Car-NeRF model can serve as a component for an editable autonomous driving simulator. As the scale of CarPatch3D expands, we anticipate that our model's performance in completely invisible perspective synthesis and appearance editing will continue to improve.







\bibliographystyle{IEEEtran}
\bibliography{IEEEexample}

\begin{thebibliography}{10}
\providecommand{\url}[1]{#1}
\csname url@rmstyle\endcsname
\providecommand{\newblock}{\relax}
\providecommand{\bibinfo}[2]{#2}
\providecommand\BIBentrySTDinterwordspacing{\spaceskip=0pt\relax}
\providecommand\BIBentryALTinterwordstretchfactor{4}
\providecommand\BIBentryALTinterwordspacing{\spaceskip=\fontdimen2\font plus
\BIBentryALTinterwordstretchfactor\fontdimen3\font minus
  \fontdimen4\font\relax}
\providecommand\BIBforeignlanguage[2]{{%
\expandafter\ifx\csname l@#1\endcsname\relax
\typeout{** WARNING: IEEEtran.bst: No hyphenation pattern has been}%
\typeout{** loaded for the language `#1'. Using the pattern for}%
\typeout{** the default language instead.}%
\else
\language=\csname l@#1\endcsname
\fi
#2}}

\bibitem{wu2019detectron2}
Y.~Wu, A.~Kirillov, F.~Massa, W.-Y. Lo, and R.~Girshick, ``Detectron2,''
  \url{https://github.com/facebookresearch/detectron2}, 2019.

\bibitem{dd3d}
D.~Park, R.~Ambrus, V.~C. Guizilini, J.~Li, and A.~Gaidon, ``Is pseudo-lidar
  needed for monocular 3d object detection?'' \emph{2021 IEEE/CVF International
  Conference on Computer Vision (ICCV)}, pp. 3122--3132, 2021.

\bibitem{SAM}
A.~Kirillov, E.~Mintun, N.~Ravi, H.~Mao, C.~Rolland, L.~Gustafson, T.~Xiao,
  S.~Whitehead, A.~C. Berg, W.-Y. Lo, P.~Doll{\'a}r, and R.~B. Girshick,
  ``Segment anything,'' \emph{ArXiv}, vol. abs/2304.02643, 2023.

\bibitem{Nerf}
B.~Mildenhall, P.~P. Srinivasan, M.~Tancik, J.~T. Barron, R.~Ramamoorthi, and
  R.~Ng, ``Nerf: Representing scenes as neural radiance fields for view
  synthesis,'' \emph{ArXiv}, vol. abs/2003.08934, 2020.

\bibitem{autorf}
N.~M{\"u}ller, A.~Simonelli, L.~Porzi, S.~R. Bul{\`o}, M.~Nie{\ss}ner, and
  P.~Kontschieder, ``Autorf: Learning 3d object radiance fields from single
  view observations,'' \emph{2022 IEEE/CVF Conference on Computer Vision and
  Pattern Recognition (CVPR)}, pp. 3961--3970, 2022.

\bibitem{snerf}
Z.~Xie, J.~Zhang, W.~Li, F.~Zhang, and L.~Zhang, ``S-nerf: Neural radiance
  fields for street views,'' \emph{ArXiv}, vol. abs/2303.00749, 2023.

\bibitem{PNF}
A.~Kundu, K.~Genova, X.~Yin, A.~Fathi, C.~Pantofaru, L.~J. Guibas,
  A.~Tagliasacchi, F.~Dellaert, and T.~A. Funkhouser, ``Panoptic neural fields:
  A semantic object-aware neural scene representation,'' \emph{2022 IEEE/CVF
  Conference on Computer Vision and Pattern Recognition (CVPR)}, pp.
  12\,861--12\,871, 2022.

\bibitem{SUDS}
H.~Turki, J.~Y. Zhang, F.~Ferroni, and D.~Ramanan, ``Suds: Scalable urban
  dynamic scenes,'' \emph{ArXiv}, vol. abs/2303.14536, 2023.

\bibitem{discoscene}
Y.~Xu, M.~Chai, Z.~Shi, S.~Peng, I.~Skorokhodov, A.~Siarohin, C.~Yang, Y.~Shen,
  H.-Y. Lee, B.~Zhou, and S.~Tulyakov, ``Discoscene: Spatially disentangled
  generative radiance fields for controllable 3d-aware scene synthesis,''
  \emph{ArXiv}, vol. abs/2212.11984, 2022.

\bibitem{neuralscenegraphs}
J.~Ost, F.~Mannan, N.~Thuerey, J.~Knodt, and F.~Heide, ``Neural scene graphs
  for dynamic scenes,'' \emph{2021 IEEE/CVF Conference on Computer Vision and
  Pattern Recognition (CVPR)}, pp. 2855--2864, 2020.

\bibitem{gina3d}
B.~Shen, X.~Yan, C.~Qi, M.~Najibi, B.~Deng, L.~J. Guibas, Y.~Zhou, and
  D.~Anguelov, ``Gina-3d: Learning to generate implicit neural assets in the
  wild,'' \emph{ArXiv}, vol. abs/2304.02163, 2023.

\bibitem{geosim}
Y.~Chen, F.~Rong, S.~Duggal, S.~Wang, X.~Yan, S.~Manivasagam, S.~Xue, E.~Yumer,
  and R.~Urtasun, ``Geosim: Realistic video simulation via geometry-aware
  composition for self-driving,'' \emph{2021 IEEE/CVF Conference on Computer
  Vision and Pattern Recognition (CVPR)}, pp. 7226--7236, 2021.

\bibitem{ners}
J.~Y. Zhang, G.~Yang, S.~Tulsiani, and D.~Ramanan, ``Ners: Neural reflectance
  surfaces for sparse-view 3d reconstruction in the wild,'' in \emph{Neural
  Information Processing Systems}, 2021.

\bibitem{StreetSurf}
J.~Guo, N.~Deng, X.~Li, Y.~Bai, B.~Shi, C.~Wang, C.~Ding, D.~Wang, and Y.~Li,
  ``Streetsurf: Extending multi-view implicit surface reconstruction to street
  views,'' \emph{ArXiv}, vol. abs/2306.04988, 2023.

\bibitem{BlockNeRF}
M.~Tancik, V.~Casser, X.~Yan, S.~Pradhan, B.~Mildenhall, P.~P. Srinivasan,
  J.~T. Barron, and H.~Kretzschmar, ``Block-nerf: Scalable large scene neural
  view synthesis,'' \emph{2022 IEEE/CVF Conference on Computer Vision and
  Pattern Recognition (CVPR)}, pp. 8238--8248, 2022.

\bibitem{FEGR}
Z.~Wang, T.~Shen, J.~Gao, S.~Y. Huang, J.~Munkberg, J.~Hasselgren, Z.~Gojcic,
  W.~Chen, and S.~Fidler, ``Neural fields meet explicit geometric
  representation for inverse rendering of urban scenes,'' \emph{ArXiv}, vol.
  abs/2304.03266, 2023.

\bibitem{urbangiraffe}
Y.~Yang, Y.~Yang, H.~Guo, R.~Xiong, Y.~Wang, and Y.~Liao, ``Urbangiraffe:
  Representing urban scenes as compositional generative neural feature
  fields,'' \emph{ArXiv}, vol. abs/2303.14167, 2023.

\bibitem{airsim}
S.~Shah, D.~Dey, C.~Lovett, and A.~Kapoor, ``Airsim: High-fidelity visual and
  physical simulation for autonomous vehicles,'' in \emph{International
  Symposium on Field and Service Robotics}, 2017.

\bibitem{carla}
A.~Dosovitskiy, G.~Ros, F.~Codevilla, A.~M. L{\'o}pez, and V.~Koltun, ``Carla:
  An open urban driving simulator,'' \emph{ArXiv}, vol. abs/1711.03938, 2017.

\bibitem{kitti}
A.~Geiger, P.~Lenz, and R.~Urtasun, ``Are we ready for autonomous driving? the
  kitti vision benchmark suite,'' \emph{2012 IEEE Conference on Computer Vision
  and Pattern Recognition}, pp. 3354--3361, 2012.

\bibitem{dvm}
J.~Huang, B.~Chen, L.~Luo, S.~Yue, and I.~Ounis, ``Dvm-car: A large-scale
  automotive dataset for visual marketing research and applications,''
  \emph{2022 IEEE International Conference on Big Data (Big Data)}, pp.
  4140--4147, 2021.

\bibitem{mipnerf}
J.~T. Barron, B.~Mildenhall, M.~Tancik, P.~Hedman, R.~Martin-Brualla, and P.~P.
  Srinivasan, ``Mip-nerf: A multiscale representation for anti-aliasing neural
  radiance fields,'' \emph{2021 IEEE/CVF International Conference on Computer
  Vision (ICCV)}, pp. 5835--5844, 2021.

\bibitem{DynamicViewSynthesis}
C.~Gao, A.~Saraf, J.~Kopf, and J.-B. Huang, ``Dynamic view synthesis from
  dynamic monocular video,'' \emph{2021 IEEE/CVF International Conference on
  Computer Vision (ICCV)}, pp. 5692--5701, 2021.

\bibitem{nerfdiff}
J.~Gu, A.~Trevithick, K.-E. Lin, J.~M. Susskind, C.~Theobalt, L.~Liu, and
  R.~Ramamoorthi, ``Nerfdiff: Single-image view synthesis with nerf-guided
  distillation from 3d-aware diffusion,'' \emph{ArXiv}, vol. abs/2302.10109,
  2023.

\bibitem{3DiM}
D.~Watson, W.~Chan, R.~Martin-Brualla, J.~Ho, A.~Tagliasacchi, and M.~Norouzi,
  ``Novel view synthesis with diffusion models,'' \emph{ArXiv}, vol.
  abs/2210.04628, 2022.

\bibitem{Zero1to3}
R.~Liu, R.~Wu, B.~V. Hoorick, P.~Tokmakov, S.~Zakharov, and C.~Vondrick,
  ``Zero-1-to-3: Zero-shot one image to 3d object,'' \emph{ArXiv}, vol.
  abs/2303.11328, 2023.

\bibitem{deceptivenerf}
X.~Liu, S.~hong Kao, J.~Chen, Y.-W. Tai, and C.-K. Tang, ``Deceptive-nerf:
  Enhancing nerf reconstruction using pseudo-observations from diffusion
  models,'' \emph{ArXiv}, vol. abs/2305.15171, 2023.

\bibitem{Magic123}
G.~Qian, J.~Mai, A.~Hamdi, J.~Ren, A.~Siarohin, B.~Li, H.-Y. Lee,
  I.~Skorokhodov, P.~Wonka, S.~Tulyakov, and B.~Ghanem, ``Magic123: One image
  to high-quality 3d object generation using both 2d and 3d diffusion priors,''
  \emph{ArXiv}, vol. abs/2306.17843, 2023.

\bibitem{PixelNeRF}
A.~Yu, V.~Ye, M.~Tancik, and A.~Kanazawa, ``pixelnerf: Neural radiance fields
  from one or few images,'' \emph{2021 IEEE/CVF Conference on Computer Vision
  and Pattern Recognition (CVPR)}, pp. 4576--4585, 2020.

\bibitem{CMR}
A.~Kanazawa, S.~Tulsiani, A.~A. Efros, and J.~Malik, ``Learning
  category-specific mesh reconstruction from image collections,'' \emph{ArXiv},
  vol. abs/1803.07549, 2018.

\bibitem{pixel2mesh}
N.~Wang, Y.~Zhang, Z.~Li, Y.~Fu, W.~Liu, and Y.-G. Jiang, ``Pixel2mesh:
  Generating 3d mesh models from single rgb images,'' in \emph{European
  Conference on Computer Vision}, 2018.

\bibitem{deformation}
P.~Henderson, V.~Tsiminaki, and C.~H. Lampert, ``Leveraging 2d data to learn
  textured 3d mesh generation,'' \emph{2020 IEEE/CVF Conference on Computer
  Vision and Pattern Recognition (CVPR)}, pp. 7495--7504, 2020.

\bibitem{nvdiffrec}
J.~Munkberg, J.~Hasselgren, T.~Shen, J.~Gao, W.~Chen, A.~Evans, T.~M{\"u}ller,
  and S.~Fidler, ``Extracting triangular 3d models, materials, and lighting
  from images,'' \emph{2022 IEEE/CVF Conference on Computer Vision and Pattern
  Recognition (CVPR)}, pp. 8270--8280, 2021.

\bibitem{unsup3d}
S.~Wu, C.~Rupprecht, and A.~Vedaldi, ``Unsupervised learning of probably
  symmetric deformable 3d objects from images in the wild,'' \emph{2020
  IEEE/CVF Conference on Computer Vision and Pattern Recognition (CVPR)}, pp.
  1--10, 2019.

\bibitem{meshrcnn}
G.~Gkioxari, J.~Malik, and J.~Johnson, ``Mesh r-cnn,'' \emph{2019 IEEE/CVF
  International Conference on Computer Vision (ICCV)}, pp. 9784--9794, 2019.

\bibitem{pifu}
S.~Saito, Z.~Huang, R.~Natsume, S.~Morishima, A.~Kanazawa, and H.~Li, ``Pifu:
  Pixel-aligned implicit function for high-resolution clothed human
  digitization,'' \emph{2019 IEEE/CVF International Conference on Computer
  Vision (ICCV)}, pp. 2304--2314, 2019.

\bibitem{3D-RCNN}
A.~Kundu, Y.~Li, and J.~M. Rehg, ``3d-rcnn: Instance-level 3d object
  reconstruction via render-and-compare,'' \emph{2018 IEEE/CVF Conference on
  Computer Vision and Pattern Recognition}, pp. 3559--3568, 2018.

\bibitem{CADSim}
J.~Wang, S.~Manivasagam, Y.~Chen, Z.~Yang, I.~A. B{\^a}rsan, A.~Yang, W.-C. Ma,
  and R.~Urtasun, ``Cadsim: Robust and scalable in-the-wild 3d reconstruction
  for controllable sensor simulation,'' in \emph{Conference on Robot Learning},
  2022.

\bibitem{singleshotsr}
S.~Zakharov, R.~Ambrus, V.~C. Guizilini, D.~Park, W.~Kehl, F.~Durand, J.~B.
  Tenenbaum, V.~Sitzmann, J.~Wu, and A.~Gaidon, ``Single-shot scene
  reconstruction,'' in \emph{Conference on Robot Learning}, 2021.

\bibitem{ObjectCompositinalNeRF}
B.~Yang, Y.~Zhang, Y.~Xu, Y.~Li, H.~Zhou, H.~Bao, G.~Zhang, and Z.~Cui,
  ``Learning object-compositional neural radiance field for editable scene
  rendering,'' \emph{2021 IEEE/CVF International Conference on Computer Vision
  (ICCV)}, pp. 13\,759--13\,768, 2021.

\bibitem{MVSNerf}
A.~Chen, Z.~Xu, F.~Zhao, X.~Zhang, F.~Xiang, J.~Yu, and H.~Su, ``Mvsnerf: Fast
  generalizable radiance field reconstruction from multi-view stereo,''
  \emph{2021 IEEE/CVF International Conference on Computer Vision (ICCV)}, pp.
  14\,104--14\,113, 2021.

\bibitem{visionNerf}
K.-E. Lin, Y.-C. Lin, W.-S. Lai, T.-Y. Lin, Y.~Shih, and R.~Ramamoorthi,
  ``Vision transformer for nerf-based view synthesis from a single input
  image,'' \emph{2023 IEEE/CVF Winter Conference on Applications of Computer
  Vision (WACV)}, pp. 806--815, 2022.

\bibitem{SRT}
M.~S.~M. Sajjadi, H.~Meyer, E.~Pot, U.~M. Bergmann, K.~Greff, N.~Radwan,
  S.~Vora, M.~Lucic, D.~Duckworth, A.~Dosovitskiy, J.~Uszkoreit, T.~A.
  Funkhouser, and A.~Tagliasacchi, ``Scene representation transformer:
  Geometry-free novel view synthesis through set-latent scene
  representations,'' \emph{2022 IEEE/CVF Conference on Computer Vision and
  Pattern Recognition (CVPR)}, pp. 6219--6228, 2021.

\bibitem{codenerf}
W.~J. Jang and L.~de~Agapito, ``Codenerf: Disentangled neural radiance fields
  for object categories,'' \emph{2021 IEEE/CVF International Conference on
  Computer Vision (ICCV)}, pp. 12\,929--12\,938, 2021.

\bibitem{resnet}
K.~He, X.~Zhang, S.~Ren, and J.~Sun, ``Deep residual learning for image
  recognition,'' \emph{2016 IEEE Conference on Computer Vision and Pattern
  Recognition (CVPR)}, pp. 770--778, 2015.

\bibitem{imagenet}
J.~Deng, W.~Dong, R.~Socher, L.-J. Li, K.~Li, and L.~Fei-Fei, ``Imagenet: A
  large-scale hierarchical image database,'' \emph{2009 IEEE Conference on
  Computer Vision and Pattern Recognition}, pp. 248--255, 2009.

\bibitem{nerfstudio}
M.~Tancik, E.~Weber, E.~Ng, R.~Li, B.~Yi, J.~Kerr, T.~Wang, A.~Kristoffersen,
  J.~Austin, K.~Salahi, A.~Ahuja, D.~McAllister, and A.~Kanazawa, ``Nerfstudio:
  A modular framework for neural radiance field development,'' \emph{ArXiv},
  vol. abs/2302.04264, 2023.

\bibitem{autorf-pytorch}
``Autorf-pytorch,'' \url{https://github.com/skyhehe123/AutoRF-pytorch}.

\bibitem{radam}
L.~Liu, H.~Jiang, P.~He, W.~Chen, X.~Liu, J.~Gao, and J.~Han, ``On the variance
  of the adaptive learning rate and beyond,'' \emph{ArXiv}, vol.
  abs/1908.03265, 2019.

\bibitem{ssim}
Z.~Wang, E.~P. Simoncelli, and A.~C. Bovik, ``Multiscale structural similarity
  for image quality assessment,'' \emph{The Thrity-Seventh Asilomar Conference
  on Signals, Systems \& Computers, 2003}, vol.~2, pp. 1398--1402 Vol.2, 2003.

\bibitem{lpips}
R.~Zhang, P.~Isola, A.~A. Efros, E.~Shechtman, and O.~Wang, ``The unreasonable
  effectiveness of deep features as a perceptual metric,'' \emph{2018 IEEE/CVF
  Conference on Computer Vision and Pattern Recognition}, pp. 586--595, 2018.

\end{thebibliography}

\end{document}